\documentclass[11pt]{article}

\usepackage[preprint]{acl}

\usepackage{times}
\usepackage{latexsym}

\usepackage[T1]{fontenc}

\usepackage[utf8]{inputenc}

\usepackage{microtype}

\usepackage{inconsolata}

\usepackage{graphicx}

\usepackage{xspace}
\usepackage{amsmath}
\usepackage{booktabs}
\usepackage{amssymb}
\usepackage{array}
\usepackage{enumitem}
\usepackage{float}
\usepackage{listings}
\definecolor{darkcyan}{rgb}{0.0, 0.55, 0.55}

\newcommand{\sys}{CIRCE\xspace}
\newcommand{\syss}{CIRCE-Stats\xspace}
\newcommand{\sysg}{CIRCE-GNN\xspace}
\newcommand{\sysgr}{CIRCE-GNN-R\xspace}
\newcommand{\sysgp}{CIRCE-GNN-P\xspace}

\title{CIRCE Divines LLM Task Complexity from Constraint Structure}

\author{
    \textbf{Chenhao Hong} \quad
    \textbf{Jiakai Xu} \quad
    \textbf{Kaushiki Singh} \quad
    \textbf{Yusen Zhang} \quad
    \textbf{Eugene Wu} \\
    Columbia University \\
    \texttt{\{chenhao.hong,ax2155,ks4402,yz5296\}@columbia.edu} \\
    \texttt{ewu@cs.columbia.edu}
}

\begin{document}
\maketitle
\begin{abstract}
Estimating whether a problem is too complex for a large language model (LLM) before the model attempts it can guide model routing, how to allocate test-time compute, and suggest the need for new tools to aid the agent. While existing estimators rely on the problem text, LLM-annotated task difficulty labels, or signals obtained after a model attempts to solve the problem, such as its hidden states, none explicitly leverage the constraint structure inherent in the problem: the variables to assign and how the constraints over them interact. 

We introduce \sys,\footnote{Code will be released at
\url{https://github.com/hchenhao/CIRCE}.} which extracts a problem's structure by using a cheap LLM to construct a {\it Constraint Intermediate Representation} that deterministically compiles into the constraint graph.
Across seven reasoning benchmarks (including SAT, logic, and math) and four LLMs, even a randomly initialized GNN encoder improves predictions of LLM response correctness, reducing negative log-likelihood from 0.465 to 0.423 and increasing AUROC from 0.691 to 0.727 relative to the strongest baseline for each metric.

\end{abstract}

\section{Introduction}
\label{sec:introduction}

Large language models (LLMs) are increasingly asked to solve complex problems in which constraints create dependencies among decisions. These problems include constraint satisfaction (e.g., arranging wedding seating while keeping some guests together and others apart), resource management and scheduling (e.g., assigning staff to shifts while respecting availability and coverage requirements), logical entailment (e.g., determining whether an expense qualifies for reimbursement under a company’s policy), and database editing under integrity constraints (e.g., deleting a customer record without leaving orders linked to a nonexistent customer). Although task success may be assessed from an LLM's result, it remains difficult to predict in advance whether a particular problem is too complex for the model to solve reliably.

Complexity estimation predicts the probability that a given LLM will answer a problem correctly under a fixed inference protocol (prompting and generation settings). Estimates made before generation can support model routing \citep{chen2023frugalgpt,ong2025routellm}, allocation of test-time compute \citep{snell2024scalingllmtesttimecompute}, and abstention on likely failures \citep{kamath-etal-2020-selective}.

Existing approaches use text embeddings of the problem \citep{pacchiardi-etal-2025-predictaboard}, LLM-generated ratings of required abilities such as logical reasoning and domain knowledge \citep{zhou2026general}, or the target model's internal activations \citep{zhu-etal-2025-llm}. The latter requires internal model access unavailable in proprietary model APIs. These approaches also do not explicitly represent the problem's variables and interacting constraints.

Constraint structure helps predict computational difficulty: random 3-SAT formulas are hardest near a critical clause-to-variable ratio \citep{mitchell1992hard}, and SATzilla uses structural features to predict runtimes and select the solver expected to be fastest \citep{xu2008satzilla}. Constraint structure also guides LLM task decomposition \citep{zhou2025aconic}.

We therefore ask: \textit{(1) does constraint structure predict LLM response correctness, and (2) can it be recovered reliably from problem text?}

One challenge is that natural-language problems often do not specify variables and constraints explicitly, and LLM-generated constraint graphs are unreliable: only 136 of 200 Hamiltonian path graphs are correct in our evaluation (68.0\%; Table~\ref{tab:cir-rule-validation}). Our key observation is that an LLM need only \textit{describe} the variables, constraints, and query in an intermediate representation, which a deterministic compiler converts into a graph.

We introduce \textit{Constraint Intermediate Representation for Complexity Estimation} (\textbf{\sys}), a workflow built on this separation of semantic extraction and graph construction. On the Hamiltonian path, this approach raises correctness to 92.5\%. We develop 41 handcrafted graph features (\syss) and find that richer feature sets generally improve prediction. To capture interactions beyond aggregate statistics, we also use a graph neural network (\sysg). We compare frozen encoders with random initialization or self-supervised pretraining on separate unlabeled problems. A lightweight, model-specific predictor maps either representation to the probability of a correct response.

We evaluate \sys against four prediction baselines and a constant predictor on seven reasoning benchmarks and four LLMs. Five-fold cross-validation over 200 problems per benchmark fits each predictor on approximately 160 training problems, with ten responses per problem supplying correctness labels. Relative to the strongest baseline overall for each metric, the pretrained variant reduces average negative log-likelihood of response correctness by 9.5\% (0.465 to 0.421) and increases AUROC by 3.9 percentage points (0.691 to 0.730). Even a frozen, randomly initialized GNN encoder outperforms every baseline on average and nearly matches the pretrained variant. These results show that constraint structure recovered from text is a useful signal for predicting LLM failures before generation. Text-to-SQL and program synthesis are promising targets for future evaluation.

\section{Related Work}

\subsection{Complexity Estimation for LLMs}

Recent work predicts instance-level LLM performance using textual assessors \cite{pacchiardi-etal-2025-predictaboard, 10.1145/3706118}, semantic task-demand annotations \cite{zhou2026general}, or the target model's hidden representations \cite{zhu-etal-2025-llm}. Complementary work estimates difficulty from problem content or answer plausibility \cite{pmlr-v267-truong25c, mozafari-etal-2026-question}, or uses difficulty estimates to adapt reasoning strategies \cite{xu-etal-2024-adaption}. A related line of work estimates response correctness using elicited confidence, inter-model response agreement, or historical correctness patterns \cite{tian-etal-2023-just, xia-etal-2025-influences, pmlr-v306-xiao26l}. Although these approaches provide useful textual, semantic, representational, and behavioral signals, they do not explicitly model reasoning problems as constraint structures with interacting variables and constraints.

\subsection{Structured Representations for LLM Reasoning}

Prior work has explored symbolic representations for LLM reasoning \cite{pan-etal-2023-logic, olausson-etal-2023-linc},  relationships between problem structure and reasoning complexity \cite{kang-etal-2024-empirical}, and limitations in LLM reasoning over complex data structures \cite{pmlr-v306-he26x}. Related work examines formal specification inference and reliable theorem proving \cite{lecong2025llmsreasonprogramsemantics, quan2025faithfulrobustllmdriventheorem}. Other studies examine compositional problem complexity \cite{xie2026rethinkingcodecomplexitylens} and constraint-based problem decomposition \cite{zhou2025aconic}. Research on multi-constraint instruction following further shows that constraint composition and dependencies affect model performance \cite{wen2024neurips-benchmarking, liu2026adaplanbenchevaluatingadaptiveplanning}. While these studies show that problem structure and constraints matter for LLM reasoning, whether structural features extracted from a common constraint representation can predict correctness on individual problem instances remains underexplored.

\subsection{Learning from Graph Representations}

Studies show that LLMs struggle with structured graph reasoning problems, motivating explicit representations of relationships and dependencies \cite{dai2025revisitinggraphreasoningability, agrawal2024llmsperformstructuredgraph}. Structural properties have also been used to predict solver performance through handcrafted features \cite{xu2008satzilla} and graph-based learning \cite{chang2022predicting, pmlr-v258-fu25b}. Graph neural networks, in contrast, learn representations through neighborhood aggregation \cite{kipf2017semisupervisedclassificationgraphconvolutional, xu2019powerfulgraphneuralnetworks}, including for language reasoning over knowledge graphs \cite{yasunaga-etal-2021-qa}. Self-supervised methods such as GraphMAE \cite{hou2022graphmaeselfsupervisedmaskedgraph} further enable graph representation learning without problem-specific labels. However, rather than using these representations to solve problems or support reasoning directly, we leverage the constraint structure to estimate the empirical difficulty of individual problems for a target LLM.

\section{Method}
\label{sec:method}
The way constraints connect variables can affect the reasoning
required to answer a problem. We use these relationships as
a signal for complexity estimation. Extracting them directly from the problem using an LLM is challenging.  %
\sys addresses this by
asking the LLM to describe the variables and constraints in an intermediate representation (CIR), from which a rule-based compiler constructs the graph.

\begin{figure*}[t]
\centering
\includegraphics[width=\textwidth]{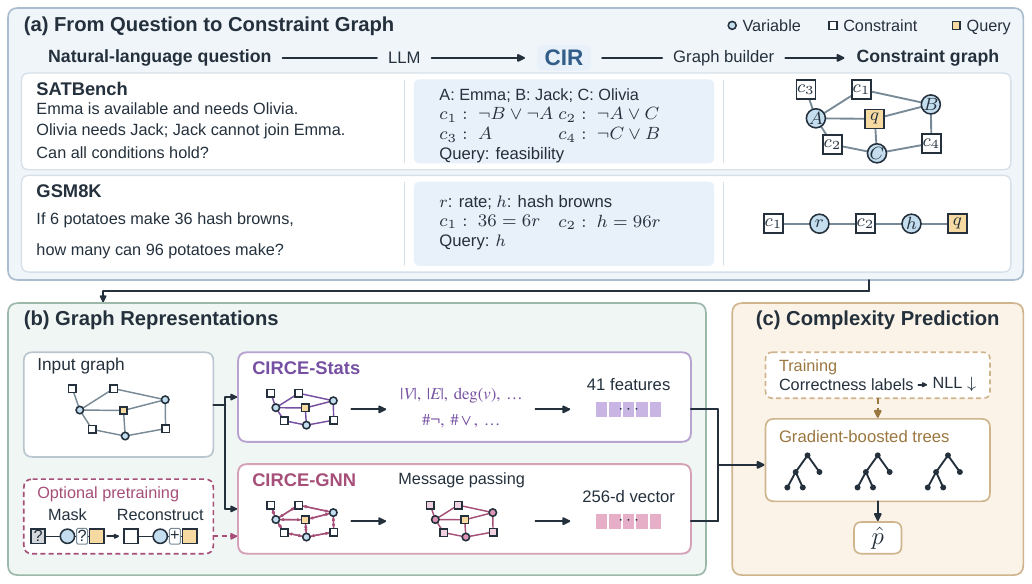}
\caption{CIRCE workflow. (a) An LLM extracts a CIR from the question, and a rule-based graph builder constructs its constraint graph. Examples from SATBench and GSM8K are abbreviated. (b) CIRCE-Stats uses handcrafted features. CIRCE-GNN produces a feature vector with GNN. Optional pretraining of the GNN masks and reconstructs node and edge attributes. (c) A gradient-boosted predictor maps the representation to an estimated score.}
\label{fig:circe}
\end{figure*}

Figure~\ref{fig:circe} illustrates the workflow of \sys.
We first extract a CIR and compile it into a constraint graph
(Section~\ref{subsec:question-to-graph}). We then represent the
graph using handcrafted features in \syss
(Section~\ref{subsec:circe-stats}) or a GNN with or without
self-supervised pretraining in \sysg
(Section~\ref{subsec:circe-gnn}). Finally, a gradient-boosted
predictor maps the structural representation to a correctness
probability for the target model
(Section~\ref{subsec:correctness-estimation}).
\subsection{Problem Formulation}
Given a problem $x$, a language model $m$, and an inference protocol $\pi$, we measure how complex $x$ is for $m$ by the probability that a response is correct,
\begin{equation}
  p_{m,\pi}(x)=\Pr(\text{a response to }x\text{ is correct}\mid m,\pi),
\end{equation}
where lower means that $x$ is more complex for $m$.
To obtain enough labels for training and evaluation, we observe $R=10$ responses per problem for each model, using a fixed inference protocol. With $c_r\in\{0,1\}$ indicating whether a response is correct, the empirical correctness rate is
\begin{equation}
  y=\sum_{r=1}^{R} c_r / R.
\end{equation}

\subsection{From Problem to Constraint Graph}
\label{subsec:question-to-graph}
A CIR describes variables, constraints, types, domains, and the requested query. We prompt an LLM to translate the problem into this representation without solving it or asserting an answer. A rule-based graph builder checks the CIR format, reports parsing errors for correction, and compiles accepted outputs into a typed constraint graph $G=(V,E)$. 

There are two types of graph nodes. \textbf{Factor nodes} represent constraints and the query, while \textbf{variable nodes} represent declared variables, propositions, entities, and shared domain values. Edges connect factor nodes to their corresponding variable nodes. For example, $50B{=}W$ creates a factor connected to bag count $B$ and weight $W$; $C{=}18B$ creates another connected to cost $C$ and $B$. The query factor connects to the requested cost $C$.

The same graph builder is shared across benchmarks. Both syntax checking and semantic validation help assess whether the CIR preserves the problem's variables, constraints, and query. Automated checks can trigger repair. Appendix~\ref{app:cir} provides the checks, repairs, and validation details.

\subsection{\syss: Handcrafted  Features}
\label{subsec:circe-stats}

Structural features have been used to predict the difficulty
of constraint-solving problems, as exemplified by
SATzilla~\cite{xu2008satzilla}. As a first step of examining whether such features can be used to predict the correctness of LLMs on constraint problems, we introduce \syss, where we handcrafted 41 features to capture the structural information of the constraint graph: size (5), factor occurrences (4), degree (8), topology (5), query (3), and CIR operators (16). Appendix~\ref{app:stats} lists the features and their definitions.

\subsection{\sysg: GNN-based Features}
\label{subsec:circe-gnn}
Experiments in Section~\ref{subsec:feature-count} show that writing more handcrafted features continued to improve the model's predictive accuracy. Yet, manual feature engineering may omit complex but predictive interactions between the constraints and the variables.
We therefore use a GNN to encode the graph structure. We project each node's attributes to an initial vector $h_v^{(0)}$ and use six GIN-style layers~\cite{xu2019powerfulgraphneuralnetworks}. Each node vector $h_v^{(\ell)}$ and edge embedding $e_{uv}^{(\ell)}$ has 64 dimensions. At layer $\ell\in\{1,\ldots,6\}$, $e_{uv}^{(\ell)}$ is a trainable embedding, and $m_v^{(\ell)}$ sums incoming neighbor messages. $\mathcal N(v)$ denotes the nodes connected to $v$, $\epsilon_\ell$ is a learned scalar, $\operatorname{MLP}_{\ell}$ a feedforward network, and $\operatorname{LN}$ layer normalization. The update is
\begin{align}
  m_v^{(\ell)}&=\sum_{u\in\mathcal{N}(v)}
  \operatorname{ReLU}\!\left(h_u^{(\ell-1)}+e_{uv}^{(\ell)}\right),\\
  h_v^{(\ell)}&=\operatorname{LN}\Big(h_v^{(\ell-1)}\notag\\
  &\quad+\operatorname{MLP}_{\ell}\!\left((1+\epsilon_{\ell})h_v^{(\ell-1)}
  +m_v^{(\ell)}\right)\Big).
\end{align}

A randomly initialized GNN is already a powerful tool in encoding graph features through its message-passing mechanism, while self-supervised pretraining may further enhance its accuracy~\cite{hou2023graphmae2decodingenhancedmaskedselfsupervised}. Thus, \sysgp uses encoder pretraining, and \sysgr skips pretraining and uses the randomly initialized parameters. \sysgp trains the GNN in a self-supervised manner by masking a portion of the node attributes and the edge attributes~\cite{hou2022graphmaeselfsupervisedmaskedgraph}. The decoders predict six attributes: node kind, node subtype, factor type, operator counts, edge role and edge polarity. Let $\mathcal A$ denote these six attributes and $\mathcal L_a$ the average reconstruction loss for attribute $a$. Node decoders receive $h_v^{(6)}$, and edge decoders receive $[h_u^{(6)};h_v^{(6)}]$. The pretraining objective is
\begin{equation}
  \mathcal L_{\mathrm{pretrain}}=\frac{1}{6}\sum_{a\in\mathcal A}\mathcal L_a.
\end{equation}

We freeze the GNN in both encoder variants after 100 pretraining epochs for \sysgp, and right after initialization for \sysgr. Let $q_v=1$ for a query node and $0$ otherwise. We concatenate the coordinate-wise sum, mean, query sum and standard deviation of the final node vectors into the graph representation
\begin{equation}
\begin{aligned}
z(G)=\bigl[&
\sum_v h_v^{(6)};\operatorname{mean}_v h_v^{(6)};\\
&\sum_v q_v h_v^{(6)};\operatorname{std}_v h_v^{(6)}
\bigr].
\end{aligned}
\end{equation}

\subsection{Complexity Estimation}
\label{subsec:correctness-estimation}
Both \syss and \sysg give a structural representation $u(x)$ without using correctness labels. We then fit a predictor $\hat p_{m,\pi}(x)=f_{m,\pi}(u(x))$ to predict the correctness probability. We use histogram gradient-boosted trees
\begin{equation}
 f_{\mathrm{GB}}(u)=\sigma\!\left(b+\eta\sum_{t=1}^{T}
 g_t(\widetilde u)\right),
\end{equation}
where $g_t$ is a fitted regression tree, and $\widetilde u$ is the representation after training-fold preprocessing. The predictor is fitted by Bernoulli log-loss against ten-attempt correctness rates and is used for both \syss and \sysg.

\section{Experiments}
\label{sec:experiments}

\subsection{Evaluation Metrics and Baselines}
\label{subsec:evaluation-metrics-baselines}
We use grouped five-fold cross-validation for each benchmark/model pair. All methods are evaluated with three different random splits. In addition, both \sysg variants average four encoder seeds, with matched random initializations.
The primary metric is Bernoulli negative log-likelihood, defined as
\begin{equation}
 \operatorname{NLL}=-\frac1n\sum_i
 [y_i\log\hat p_i+(1-y_i)\log(1-\hat p_i)].
\end{equation}
Here $y_i=k_i/10$ is the observed success rate over ten attempts, and
$\hat p_i$ is the predicted correctness probability, clipped to $[10^{-6},1-10^{-6}]$. We also report AUROC~\cite{BRADLEY19971145}.

We compare our methods against the following baselines:
\begin{itemize}[nosep]
    \item C0: training-fold mean predictor
    \item Self probability: calibrated model self-reported probability
    \item ADeLe~\cite{zhou2026general}: a random forest predictor using 18 task-demand dimensions annotated by GPT-6 Sol, and an unguessability dimension
    \item PredictaBoard~\cite{pacchiardi-etal-2025-predictaboard}: an assessor that predicts correctness from OpenAI text embeddings
    \item Already Knows~\cite{zhu-etal-2025-llm}: a hidden-state predictor trained with temporal-difference learning
\end{itemize}

\subsection{Models and Benchmarks}
\label{subsec:models-benchmarks}
Our methods and baselines are evaluated on Qwen3.5-9B \citep{qwen2026qwen35}, Qwen3.8-27B \citep{qwen2026qwen38}, Inkling \citep{thinkingmachines2026inkling} and GPT-6 Luna \citep{openai2026luna}, representing the latest generation of open-weight and proprietary LLMs at the time of our study.

A diverse set of reasoning benchmarks with explicit or implicit constraints is used (Table~\ref{tab:benchmarks}). For each benchmark, a subset of the problems is used for evaluation, and a larger, non-overlapping subset is used for pretraining our GNN.
Qwen3.8-27B is used to generate the CIR for all the benchmarks (Appendix~\ref{app:cir}).
We set the thinking effort of the models based on the difficulty of each benchmark to prevent saturated accuracy on easier benchmarks (Appendix~\ref{app:data}). 
Among the models we evaluate, Qwen3.5-9B is the only model that we can deploy locally given the limitations of model weight access and GPU memory. It is therefore used as the only model to evaluate the Already Knows predictor. To match the behavior of the models used in the original paper, we set the thinking mode to off for Qwen3.5-9B.
\begin{table}[ht]
\centering
\footnotesize
\setlength{\tabcolsep}{2pt}
\begin{tabular}{@{}>{\raggedright\arraybackslash}p{0.48\linewidth}rrp{0.16\linewidth}@{}}
\toprule
Benchmark & Pretrain & Evaluation & Thinking \\
\midrule
SATBench \citep{wei2025satbench} & 900 & 200 & Default \\
FOLIO \citep{han2024folio} & 900 & 200 & Default \\
NLCO (KP/QKP/MIS/MVC) \citep{jiang2026nlco} & 400 & 200 & Default \\
Enigmata: Hamiltonian Path \citep{chen2025enigmata} & 900 & 200 & Low \\
Enigmata: Zebra Logic \citep{chen2025enigmata} & 900 & 200 & Low \\
GSM8K \citep{cobbe2021gsm8k} & 900 & 200 & Off \\
BBH: Logical Deduction (5/7 objects) \citep{suzgun2023bbh} & 300 & 200 & Off \\
\midrule
Total & 5,200 & 1,400 & -- \\
\bottomrule
\end{tabular}
\caption{Question counts and thinking effort settings for models other than
Qwen3.5-9B, whose thinking mode is always set to Off. Pretrain questions are used only for pretraining the GNN encoder of \sysgp. Evaluation questions are used for fitting and evaluating the predictors in all methods.}
\label{tab:benchmarks}
\end{table}

\subsection{Main Results}
\label{subsec:results}

\begin{table*}[!t]
\centering
\fontsize{9}{10.5}\selectfont
\setlength{\tabcolsep}{2pt}
\renewcommand{\arraystretch}{1.04}
\begin{tabular}{@{}l*{8}{>{\centering\arraybackslash}p{45.84pt}}@{}}
\toprule
 & \makebox[45.84pt][c]{SATBench} & \makebox[45.84pt][c]{FOLIO} & \makebox[45.84pt][c]{NLCO} & \makebox[45.84pt][c]{Hamiltonian} & \makebox[45.84pt][c]{Zebra} & \makebox[45.84pt][c]{GSM8K} & \makebox[45.84pt][c]{Logical} & \makebox[45.84pt][c]{Avg}\\
\cmidrule(l{2pt}r{2pt}){2-2}\cmidrule(l{2pt}r{2pt}){3-3}\cmidrule(l{2pt}r{2pt}){4-4}\cmidrule(l{2pt}r{2pt}){5-5}\cmidrule(l{2pt}r{2pt}){6-6}\cmidrule(l{2pt}r{2pt}){7-7}\cmidrule(l{2pt}r{2pt}){8-8}\cmidrule(l{2pt}r{2pt}){9-9}
Method & \makebox[45.84pt][c]{{\fontsize{8}{9}\selectfont NLL$\downarrow$\,/\,AUC$\uparrow$}} & \makebox[45.84pt][c]{{\fontsize{8}{9}\selectfont NLL$\downarrow$\,/\,AUC$\uparrow$}} & \makebox[45.84pt][c]{{\fontsize{8}{9}\selectfont NLL$\downarrow$\,/\,AUC$\uparrow$}} & \makebox[45.84pt][c]{{\fontsize{8}{9}\selectfont NLL$\downarrow$\,/\,AUC$\uparrow$}} & \makebox[45.84pt][c]{{\fontsize{8}{9}\selectfont NLL$\downarrow$\,/\,AUC$\uparrow$}} & \makebox[45.84pt][c]{{\fontsize{8}{9}\selectfont NLL$\downarrow$\,/\,AUC$\uparrow$}} & \makebox[45.84pt][c]{{\fontsize{8}{9}\selectfont NLL$\downarrow$\,/\,AUC$\uparrow$}} & \makebox[45.84pt][c]{{\fontsize{8}{9}\selectfont NLL$\downarrow$\,/\,AUC$\uparrow$}}\\
\midrule
\multicolumn{9}{@{}l}{\textbf{Qwen3.5-9B}}\\
\addlinespace[1pt]
C0 & .5040/.4290 & .6692/.4855 & .5838/.4506 & .6481/.4732 & .2746/.4499 & .6224/.4748 & .6904/.4791 & .5703/.4631\\
Self probability & .4999/.5228 & .6760/.4662 & .5683/.6240 & .6497/.4551 & .2693/.4896 & .6036/.6146 & .7034/.4347 & .5672/.5153\\
ADeLe & .5592/.7034 & .7044/.6295 & .4895/.7641 & .5944/.6918 & .2657/.6238 & .5947/.7087 & .6158/.7243 & .5462/.6922\\
PredictaBoard & .4517/.7093 & .6693/.4855 & .5198/.7021 & .5952/.6931 & .2701/.5956 & .6322/.5128 & .6868/.5585 & .5464/.6081\\
Already Knows & .5823/.5167 & .6796/.4975 & .6381/.5059 & .6262/.7040 & .4417/.5192 & .6276/.5458 & .6922/.4879 & .6125/.5396\\
\midrule
\syss & \textcolor{red}{\textbf{.4141}}/\textcolor{red}{\textbf{.7727}} & .6577/.6144 & .4609/.8016 & .5735/.7245 & \textcolor{red}{\textbf{.2476}}/.6914 & .5580/.7169 & .6299/.6799 & .5060/.7145\\
\sysgr & .4265/.7539 & \textcolor{red}{\textbf{.6482}}/\textcolor{red}{\textbf{.6332}} & .4569/.8064 & .5657/.7384 & .2518/\textcolor{red}{\textbf{.7013}} & .5644/.7111 & .5670/.7745 & .4972/.7313\\
\sysgp & .4289/.7521 & .6520/.6278 & \textcolor{red}{\textbf{.4544}}/\textcolor{red}{\textbf{.8085}} & \textcolor{red}{\textbf{.5630}}/\textcolor{red}{\textbf{.7415}} & .2530/.6962 & \textcolor{red}{\textbf{.5566}}/\textcolor{red}{\textbf{.7199}} & \textcolor{red}{\textbf{.5634}}/\textcolor{red}{\textbf{.7779}} & \textcolor{red}{\textbf{.4959}}/\textcolor{red}{\textbf{.7320}}\\
\midrule
\multicolumn{9}{@{}l}{\textbf{Qwen3.8-27B}}\\
\addlinespace[1pt]
C0 & .1621/.3446 & .6360/.4766 & .3581/.3927 & .3823/.4637 & .1995/.4291 & .6879/.4794 & .5828/.4756 & .4298/.4374\\
Self probability & .1671/.4923 & .6228/.6054 & .2442/.8802 & .3085/.8098 & .1934/.6412 & .6941/.5023 & .5606/.5738 & .3987/.6436\\
ADeLe & .2151/\textcolor{red}{\textbf{.6193}} & .8534/.6056 & .2700/.8508 & .3080/.8037 & .2000/.6670 & .7568/.7171 & .4618/.7786 & .4379/.7203\\
PredictaBoard & .1661/.4677 & .6557/.4873 & .2761/.8344 & .3239/.8096 & .1940/.6384 & .6680/.6004 & .5739/.5625 & .4082/.6286\\
Already Knows & -- & -- & -- & -- & -- & -- & -- & --\\
\midrule
\syss & .1619/.5685 & .6320/.6088 & \textcolor{red}{\textbf{.2389}}/\textcolor{red}{\textbf{.8822}} & .2956/.8355 & \textcolor{red}{\textbf{.1810}}/\textcolor{red}{\textbf{.7422}} & .6125/.7174 & .5052/.7387 & .3753/.7276\\
\sysgr & \textcolor{red}{\textbf{.1605}}/.5974 & .6359/.6064 & .2521/.8682 & .2918/\textcolor{red}{\textbf{.8405}} & .1847/.7195 & \textcolor{red}{\textbf{.5971}}/\textcolor{red}{\textbf{.7371}} & \textcolor{red}{\textbf{.4295}}/\textcolor{red}{\textbf{.8268}} & .3645/.7423\\
\sysgp & .1606/.6018 & \textcolor{red}{\textbf{.6221}}/\textcolor{red}{\textbf{.6295}} & .2495/.8725 & \textcolor{red}{\textbf{.2909}}/.8402 & .1826/.7339 & .6022/.7330 & .4297/.8241 & \textcolor{red}{\textbf{.3625}}/\textcolor{red}{\textbf{.7479}}\\
\midrule
\multicolumn{9}{@{}l}{\textbf{Inkling}}\\
\addlinespace[1pt]
C0 & .0864/.3069 & .6548/.4751 & .2411/.4144 & .2020/.4518 & .6934/.4815 & .6226/.4813 & .5609/.4839 & .4373/.4421\\
Self probability & .0880/.4178 & .6539/.5558 & .2124/.8111 & .1946/.6102 & .6982/.4953 & .5709/.6952 & .5463/.6965 & .4235/.6117\\
ADeLe & .1588/.4653 & .6995/.6341 & .2185/.7654 & .1783/.7848 & .7023/.6048 & .6807/.6625 & .4967/.7661 & .4478/.6690\\
PredictaBoard & .0874/.3421 & .6604/.4750 & .2100/.7563 & .1796/.8181 & .6879/.5689 & .6227/.4813 & .5625/.4712 & .4301/.5590\\
Already Knows & -- & -- & -- & -- & -- & -- & -- & --\\
\midrule
\syss & .0891/.3894 & .6551/.5961 & \textcolor{red}{\textbf{.1909}}/.8248 & \textcolor{red}{\textbf{.1575}}/.8455 & \textcolor{red}{\textbf{.6452}}/\textcolor{red}{\textbf{.6626}} & .5437/.7419 & .5338/.6550 & .4022/.6736\\
\sysgr & .0862/.5562 & .6455/.6196 & .1931/\textcolor{red}{\textbf{.8403}} & .1589/.8467 & .6531/.6583 & \textcolor{red}{\textbf{.5358}}/\textcolor{red}{\textbf{.7520}} & \textcolor{red}{\textbf{.4677}}/\textcolor{red}{\textbf{.7710}} & \textcolor{red}{\textbf{.3915}}/.7206\\
\sysgp & \textcolor{red}{\textbf{.0856}}/\textcolor{red}{\textbf{.5869}} & \textcolor{red}{\textbf{.6308}}/\textcolor{red}{\textbf{.6439}} & .1948/.8309 & .1591/\textcolor{red}{\textbf{.8471}} & .6543/.6579 & .5492/.7334 & .4713/.7653 & .3922/\textcolor{red}{\textbf{.7236}}\\
\midrule
\multicolumn{9}{@{}l}{\textbf{GPT-6 Luna}}\\
\addlinespace[1pt]
C0 & .2744/.4203 & .6369/.4822 & .3388/.4138 & .1701/.4417 & .6622/.4808 & .6320/.4836 & .6838/.4733 & .4855/.4565\\
Self probability & .2758/.4985 & .6256/.5774 & \textcolor{red}{\textbf{.2969}}/.7180 & .1562/.7198 & .6592/.5598 & .6309/.5895 & .6849/.4817 & .4756/.5921\\
ADeLe & .2802/.6175 & .8300/.5765 & .3039/\textcolor{red}{\textbf{.7410}} & .1566/.7495 & .5618/.7604 & .6785/.7234 & .7146/.6085 & .5036/.6824\\
PredictaBoard & .2651/.6460 & .6703/.5005 & .3132/.7009 & .1614/.7230 & .5945/.7366 & .6458/.5420 & .6717/.5820 & .4746/.6330\\
Already Knows & -- & -- & -- & -- & -- & -- & -- & --\\
\midrule
\syss & .2588/.6396 & .6140/.6418 & .3008/.7249 & .1550/.7393 & .5394/.7782 & .5528/.7411 & .6652/.6054 & .4409/.6958\\
\sysgr & \textcolor{red}{\textbf{.2545}}/\textcolor{red}{\textbf{.6801}} & .6380/.6044 & .3006/.7366 & \textcolor{red}{\textbf{.1543}}/\textcolor{red}{\textbf{.7523}} & .5359/.7831 & .5499/.7418 & .6307/.6863 & .4377/.7121\\
\sysgp & .2588/.6642 & \textcolor{red}{\textbf{.6022}}/\textcolor{red}{\textbf{.6672}} & .3025/.7287 & .1557/.7368 & \textcolor{red}{\textbf{.5346}}/\textcolor{red}{\textbf{.7846}} & \textcolor{red}{\textbf{.5422}}/\textcolor{red}{\textbf{.7505}} & \textcolor{red}{\textbf{.6263}}/\textcolor{red}{\textbf{.6924}} & \textcolor{red}{\textbf{.4318}}/\textcolor{red}{\textbf{.7178}}\\
\bottomrule
\end{tabular}
\caption{The NLL and AUROC (denoted as AUC) of all methods on four models across seven benchmarks. Each cell reports NLL/AUC. Avg is the mean across seven benchmarks. Lower NLL and higher AUC are better. The best scores within each model are marked in red and bold.}
\label{tab:results}
\end{table*}

Our methods, \syss and both variants of \sysg, have lower NLL and higher AUROC on average across the seven benchmarks than every baseline (Table~\ref{tab:results}). To summarize overall performance, we additionally compute macro averages, giving equal weight to all 28 benchmark/model pairs.

\syss is a strong method in its own right, achieving a macro NLL of 0.4311 and AUROC of 0.7029. Compared with the respective best baselines, it reduces NLL by 7.26\% relative to PredictaBoard (0.4648) and improves AUROC by 1.19 percentage points over ADeLe (0.6910).
Without encoder pretraining, \sysgr achieves a macro NLL of 0.4227 and AUROC of 0.7266, also outperforming every external baseline on average. Pretraining further reduces NLL by 0.51\% and improves AUROC by 0.38 percentage points relative to the random encoder. The results demonstrate the effectiveness of utilizing the structural information of the reasoning tasks through the extracted CIRs to improve the prediction of correctness. The results also support the use of GNN-encoded graph representations over the handcrafted features.

Compared with \sysgr, the pretrained variant brings the largest improvement on FOLIO, where AUROC is improved by 2.6 percentage points. The changes on the other six benchmarks are smaller. This modest improvement suggests that the graph representation and GNN architecture already provide useful features without pretraining. 
The advantage of \sysgp over \syss also varies
across benchmarks. The largest gains occur on the Logical benchmark, where \sysgp reduces NLL from 0.5835 to 0.5227 ($-$10.4\%) and increases AUROC from 0.6698 to 0.7649 ($+$9.5 percentage points), averaged over four models.

Among the three external baselines, ADeLe achieves the highest macro AUROC, whereas PredictaBoard achieves the lowest macro NLL across the four models. However, neither consistently outperforms Self probability on all models. We also observe an unstable performance of the Already Knows predictor, with its average NLL (0.6125) higher than that of C0 (0.5703) on the only model we tested, Qwen3.5-9B. We suspect that the Already Knows predictor is sensitive to the number of correctness labels: we provided only approximately 160 labels for training in each fold, compared with thousands of labels used in the original paper. The full implementation details are in Appendix~\ref{app:baselines}.

\section{Further Analysis}
\label{sec:analysis}

\subsection{Effect of Feature Count}
\label{subsec:feature-count}
\begin{figure*}[t]
\centering
\includegraphics[width=0.9\textwidth]{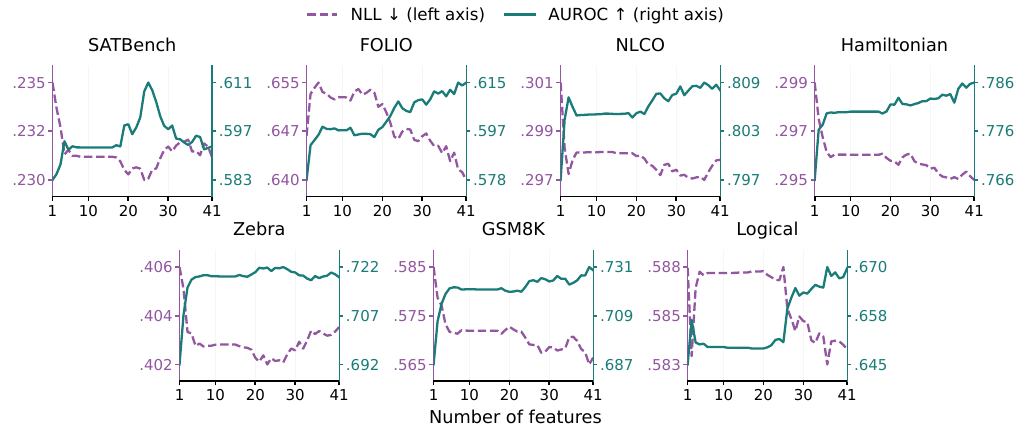}
\caption{\syss with features added one at a time using training-only forward selection. NLL (left axis) and AUROC (right axis) are averaged over four models and three splits.}
\label{fig:feature-count}
\end{figure*}

We examine how prediction changes as more handcrafted features are included. We use greedy forward selection, starting from an empty feature set. At each step, we evaluate every remaining feature together with the selected features and add the one yielding the lowest inner-validation NLL.

Figure~\ref{fig:feature-count} shows that richer feature sets generally improve prediction. Improvements continue near the full 41-feature set on several benchmarks, including FOLIO, GSM8K, and Logical.
Averaged across benchmarks and models, increasing the feature count from 30 to 41 reduces NLL from 0.4325 to 0.4311 and increases AUROC from 0.7007 to 0.7029.
These results suggest that capturing additional structural
properties can improve complexity estimation.

\subsection{Qualitative Analysis}
\begin{figure}[H]
\centering
\includegraphics[width=\columnwidth]{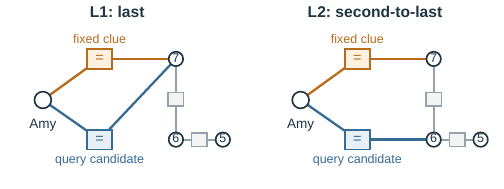}
\caption{Selected Amy-candidate connections in L1 and L2. Orange and blue squares denote fixed-position and query factors. Circles marked 5--7 are position nodes. Gray squares connect adjacent domain positions. The query shares the fixed clue's position node only in L1. Other candidates and constraints are omitted.}
\label{fig:case_relations}
\end{figure}

\paragraph{Graph representations versus statistics.}
The Logical benchmark shows the largest average advantage of \sysg over \syss. To examine what the graph representation adds, consider two problems about the same seven golfers, L1 and L2 (Figure~\ref{fig:case_relations}). Their six ordering clues are identical: Amy finishes last, Ana third-to-last, Eve third, and Eli above Dan above Ada above Mel. L1 asks who finishes last, and L2 asks who finishes second-to-last. Forty of their 41 Stats features match exactly, including their graph sizes and operator counts.

The relevant difference is how the query connects to the stated constraints. In L1, the query's position node is shared with the clue fixing Amy's position. In L2, the requested position has no such fixed-position clue. Identifying its occupant requires combining the ordering chain with the occupied positions. Message passing can distinguish these query neighborhoods, whereas the present aggregate statistics largely omit their arrangement. For Qwen3.5-9B, \sysgp predicts 0.804 and 0.282, consistent with the observed success rates of 10/10 and 0/10. \syss predicts 0.366 and 0.503, reversing their ordering. This pair illustrates a structural distinction the GNN can represent.

\paragraph{Explicit constraints versus external assessors.}
A GSM8K problem asks the cost of 850 kilograms of coal, delivered in 50-kilogram bags costing 18 dollars each. The CIR exposes the relations $W=850$, $50B=W$ and $C=18B$, where $W$ is the total weight, $B$ the number of bags and $C$ the requested cost. Its graph is a short, acyclic dependency chain: each quantity depends on one preceding quantity, without branching alternatives or additional interacting conditions. These properties provide a concrete description of the required reasoning independent of the story's wording. \syss summarizes such graph properties, while \sysg learns from the connected sequence of constraint types.

In comparison, ADeLe represents the problem through broad reasoning-demand scores, PredictaBoard uses a text embedding, and Already Knows uses the target model's initial hidden state. These inputs do not explicitly supply the extracted dependency chain: relations must be summarized by the demand scores or recovered implicitly from the learned representations. For Qwen3.5-9B, both \sys variants give this question higher probabilities than the external assessors: 0.667 for Stats and 0.737 for pretrained GNN, versus 0.253--0.419 for the three external baselines, where the target model answers all ten attempts correctly. This example illustrates a case where the explicit constraint structure is a more informative representation than the external assessors' features.

\subsection{Reliability of CIR}
We measure two metrics to assess the reliability of the CIRs: parsability and correctness. 
Parsability measures whether the CIR has a valid syntax, and correctness measures whether the CIR correctly represents the constraint structure of the problem.
For SATBench, Logical, and Hamiltonian, a rule-based checker can verify the constraint structure, so we check both parsability and correctness on all 200 problems.
For FOLIO, NLCO, Zebra, and GSM8K, rule-based checker only covers part of the constraint structure, so we check parsability on all 200 problems and manually review the correctness of 20 randomly sampled source--CIR pairs per benchmark.
On the three benchmarks with a full checker, we also check the parsability and correctness of direct graph generation, where the LLM is prompted to generate the graph directly from the natural-language question.
\begin{table}[ht]
\centering\small
\setlength{\tabcolsep}{3pt}
\begin{tabular}{lll}
\toprule
Method & Parsable & Correct\\
\midrule
\multicolumn{3}{l}{\textbf{SATBench}}\\
CIR & 198/200 = 99.0\% & 194/200 = 97.0\%\\
Direct graph & 197/200 = 98.5\% & 192/200 = 96.0\%\\
\midrule
\multicolumn{3}{l}{\textbf{Logical}}\\
CIR & 200/200 = 100.0\% & 194/200 = 97.0\%\\
Direct graph & 200/200 = 100.0\% & 185/200 = 92.5\%\\
\midrule
\multicolumn{3}{l}{\textbf{Hamiltonian}}\\
CIR & 200/200 = 100.0\% & 185/200 = 92.5\%\\
Direct graph & 196/200 = 98.0\% & 136/200 = 68.0\%\\
\bottomrule
\end{tabular}
\caption{Parsability and correctness of the CIR and direct graph generation methods on three benchmarks, with 200 questions per benchmark. The numbers are before LLM repairs.}
\label{tab:cir-rule-validation}
\end{table}

\begin{table}[ht]
\centering\small
\setlength{\tabcolsep}{3pt}
\begin{tabular}{lll}
\toprule
Benchmark & Parsable & Correct\\
\midrule
FOLIO & 200/200 = 100.0\% & 18/20 = 90.0\%\\
NLCO & 200/200 = 100.0\% & 20/20 = 100.0\%\\
Zebra & 200/200 = 100.0\% & 20/20 = 100.0\%\\
GSM8K & 200/200 = 100.0\% & 20/20 = 100.0\%\\
\bottomrule
\end{tabular}
\caption{Parsability and correctness of the CIR on four benchmarks. Parsability is checked on 200 questions per benchmark before LLM repairs, while correctness is human-reviewed on a random sample of 20 questions per benchmark after repairs.}
\label{tab:cir-spot-validation}
\end{table}

As shown in Table~\ref{tab:cir-rule-validation}, the CIR method has higher parsability and correctness than the direct graph generation method on the three benchmarks with an available correctness checker, especially on the Hamiltonian benchmark, where the correctness of the direct graph generation method is only 68.0\%, while the correctness of the CIR method is 92.5\%. Furthermore, as shown in Table~\ref{tab:cir-rule-validation} and Table~\ref{tab:cir-spot-validation}, the CIR method achieves at least 90\% correctness on all seven benchmarks, supporting that the LLM-generated CIR reliably represents the constraint structure of the problem.

Rule-based checkers can trigger CIR repair or regeneration, while human review only assesses correctness and does not trigger repairs.
Details about processing the CIRs are described in Appendix~\ref{app:cir}.

\subsection{Comparison of Predictors}

We examine whether the advantage of graph embeddings depends on the downstream predictor. Keeping the input representations and outer evaluation folds fixed, we compare gradient-boosted trees (GB), logistic regression (LR), and a multilayer perceptron (MLP). 

\begin{table}[ht]
\centering\small
\setlength{\tabcolsep}{3pt}
\begin{tabular}{lccc}
\toprule
Predictor & CIRCE-Stats & CIRCE-GNN-R & CIRCE-GNN-P \\
\midrule
GB  & .4311 / .7029 & .4227 / .7266 & .4206 / .7303 \\
LR  & .4339 / .7024 & .4209 / .7172 & .4198 / .7200 \\
MLP & .4341 / .6947 & .4206 / .7158 & .4194 / .7211 \\
\bottomrule
\end{tabular}
\caption{Effect of the downstream predictor.
Each cell reports NLL $\downarrow$ / AUROC $\uparrow$,
averaged equally across seven benchmarks and four
target models.}
\label{tab:predictor-comparison}
\end{table}

Table~\ref{tab:predictor-comparison} shows that both GNN variants achieve lower average NLL and higher average AUROC than CIRCE-Stats with all three predictors. The advantage of graph embeddings therefore persists across linear, tree-based, and neural predictors. The pretrained variant also has slightly higher AUROC and lower NLL than the random one across the three predictors. Compared with LR and MLP, GB achieves the highest average AUROC for each representation, whereas LR and MLP yield slightly lower NLL for the GNN variants.

\section{Conclusion}
\label{sec:conclusion}

\sys demonstrates that constraint structure recovered from problem text is a useful signal for predicting LLM task complexity before generation. Strong results from handcrafted statistics and randomly initialized GNNs show that this signal can be used without encoder pretraining. Graph encodings improve over statistics on average, with pretraining providing further small gains. These findings support separating task representation from model-specific prediction: a shared constraint graph can serve different target models through lightweight predictors trained on black-box labels. Extending this representation to richer constraints and queries is a promising direction for future work.

\section{Limitations}

While \sys applies to a wide range of reasoning benchmarks, its scope is limited to the problems whose constraints are part of the input. It does not cover the problems where inferring constraints is part of the reasoning process. \sys also estimates only the part of problem complexity that constraint structure reflects, while other methods that depend on the model's states could potentially reveal the failure cases such as a misread question. Combining our method with their features remains future work.

We showed the reliability of the CIRs, but they are generated by a single LLM, and their correctness cannot be fully verified: rule-based checkers cover three benchmarks, and the other four were spot-checked on 20 questions each. We also used a fixed set of problems for pretraining the GNN encoder and evaluating the predictors. 
Evaluating how the number of problems affects the encoder and predictor performance would further establish the scalability of our method.
Exploring predictor transfer across benchmarks and models could provide further evidence of the generalizability of our method.

\section*{Acknowledgments}
This work used Jetstream2~\cite{hancock2021jetstream2} at Indiana University through allocation TRA120004 from the Advanced Cyberinfrastructure Coordination Ecosystem: Services \& Support (ACCESS) program~\cite{boerner2023access}, which is supported by National Science Foundation grants \#2138259, \#2138286, \#2138307, \#2137603, and \#2138296.
This research also received DAPLab corporate support in the form of funding and/or compute from Amazon, IntellectAI, Infosys, Tidalwave, Veris, Shopify, Microsoft, Thinking Machines, Dandy, Perplexity, and Daytona. The views and conclusions presented here are those of the authors and should not be interpreted as representing the official positions of the funding organizations.

\bibliography{custom}

\appendix

\section{Data and Evaluation Protocol}
\label{app:data}
The evaluation contains 200 problems per benchmark, four target models, and ten responses per problem (56,000 responses). The 5,200 question-disjoint Pretraining (SSL) instances comprise 4,684 training and 516 validation graphs.
\begin{table}[ht]
\centering\small
\setlength{\tabcolsep}{3pt}
\begin{tabular}{lrrrr}
\toprule
Benchmark & SSL train & SSL val. & Groups & Fold train\\
\midrule
SATBench & 810 & 90 & 27 & 154--165\\
FOLIO & 810 & 90 & 152 & 158--161\\
NLCO & 364 & 36 & 200 & 160\\
Hamiltonian & 810 & 90 & 200 & 160\\
Zebra & 810 & 90 & 200 & 160\\
GSM8K & 810 & 90 & 200 & 160\\
Logical & 270 & 30 & 200 & 160\\
\bottomrule
\end{tabular}
\caption{SSL split sizes, downstream groups and outer-training-fold sizes. Each benchmark has 200 evaluation questions.}
\label{tab:app-data}
\end{table}

\paragraph{Subsets.}
We use FOLIO v2; NLCO's KP, QKP, MIS, and MVC tasks; Enigmata's Hamiltonian Path and Zebra Logic; and BBH's five- and seven-object Logical Deduction subsets~\cite{srivastava2022beyond}. 
\paragraph{Target sampling.}
Qwen3.5-9B has thinking disabled throughout. For Qwen3.8-27B and Inkling, SATBench/FOLIO/NLCO use default reasoning, Hamiltonian/Zebra use Low, and GSM8K/Logical use Off. Luna uses Medium, Low, and None for these respective groups.

Qwen and Inkling use temperature 0.7 and top-$p$ 0.95 for the five longer-format benchmarks under the 64K effective-context protocol. GSM8K and Logical use temperature 0.7, top-$p$ 0.8 and output caps of 128 and 2,048. Luna leaves temperature and top-$p$ unset in Medium/Low mode; None mode uses 0.7/1.0. Its output cap is unset. Responses are scored using final labels, choices, numerical answers or instance-specific feasibility/objective checks.
\begin{table}[ht]
\centering\small
\setlength{\tabcolsep}{3pt}
\begin{tabular}{lrrrr}
\toprule
Benchmark & 9B & 27B & Inkling & Luna\\
\midrule
SATBench & 80.00 & 96.30 & 98.35 & 92.25\\
FOLIO & 60.95 & 66.80 & 63.85 & 66.65\\
NLCO & 73.15 & 88.65 & 93.55 & 89.50\\
Hamiltonian & 65.00 & 87.25 & 94.90 & 95.95\\
Zebra & 92.20 & 95.00 & 49.10 & 37.55\\
GSM8K & 31.35 & 44.70 & 68.60 & 67.35\\
Logical & 54.00 & 73.10 & 75.15 & 57.10\\
Avg & 65.24 & 78.83 & 77.64 & 72.34\\
\bottomrule
\end{tabular}
\caption{Target response accuracy (percent) over ten attempts per question. Qwen3.5-9B has thinking disabled.}
\label{tab:app-target-accuracy}
\end{table}

\subsection{Grouped Cross-Validation}
\label{app:evaluation}
We use five outer and three inner folds, with split seeds 11, 23, and 37. SATBench groups by dimension configuration; FOLIO groups by story; other benchmarks group by instance (Table~\ref{tab:app-data}). All attempts for a question stay together. Methods share frozen folds within each benchmark/model pair. Preprocessing and hyperparameter selection use training folds only.
\paragraph{Metrics.}
Let $k_i$ be correct attempts out of ten and $p_i$ the out-of-fold prediction. NLL is Bernoulli cross-entropy against $y_i=k_i/10$, using natural logarithms and clipping probabilities to $[10^{-6},1-10^{-6}]$. Attempt-weighted AUROC is
\begin{equation}
\mathrm{AUROC}=\frac{\sum_{i,j} k_i(10-k_j)\,w(p_i,p_j)}{(\sum_i k_i)(\sum_j(10-k_j))},
\end{equation}
where $w(a,b)$ is 1 for $a>b$, $1/2$ for ties and 0 otherwise. This corresponds to ten binary outcomes per question sharing one probability.

Metrics are computed on all 200 out-of-fold predictions, then averaged over three split seeds and four encoder seeds for GNNs. Overall averages weight the 28 benchmark/model pairs equally; Already Knows covers seven Qwen3.5-9B pairs. C0 predicts the training-fold mean.
\paragraph{Feature selection.}
Stats forward selection adds the remaining feature yielding the lowest inner-validation NLL, jointly selecting tree settings within each outer training fold.

\section{CIR Extraction, Compilation and Validation}
\label{app:cir}
\paragraph{Extraction and repair.}
We use \texttt{Qwen/Qwen3.8-27B} to extract entities, variables, constraints, and a query. Prompts and input formats vary by benchmark. Parsing, reference checks, and task-specific source checks identify repair candidates. Repair inputs contain the question, candidate CIR, and diagnostic. The general format-repair prompt permits one LLM retry.

The main results use 1,400 evaluation graphs and 5,200 SSL graphs after recorded processing, fixed before predictor cross-validation. Table~\ref{tab:app-cir-repairs} counts adopted LLM repairs. All retained graphs compile.

\begin{table}[ht]
\centering\small
\setlength{\tabcolsep}{3pt}
\begin{tabular}{lrrr}
\toprule
Benchmark & Questions & Format repairs & Content repairs\\
\midrule
\multicolumn{4}{l}{\textbf{Evaluation data}}\\
SATBench & 200 & 1 & 0\\
FOLIO & 200 & 0 & 0\\
NLCO & 200 & 0 & 0\\
Hamiltonian & 200 & 0 & 15\\
Zebra & 200 & 0 & 1\\
GSM8K & 200 & 0 & 0\\
Logical & 200 & 1 & 6\\
\midrule
\multicolumn{4}{l}{\textbf{Pretraining (SSL) data}}\\
SATBench & 900 & 1 & 7\\
FOLIO & 900 & 0 & 11\\
NLCO & 400 & 0 & 4\\
Hamiltonian & 900 & 0 & 101\\
Zebra & 900 & 0 & 11\\
GSM8K & 900 & 0 & 1\\
Logical & 300 & 6 & 11\\
\bottomrule
\end{tabular}
\caption{Numbers of questions with recorded format repair, content repair based on the visible question. Format and content counts can overlap.}
\label{tab:app-cir-repairs}
\end{table}

\paragraph{Compilation.}
Factors represent constraints and the query. Non-factor nodes represent variables, propositions, entities, and shared domain values. Each occurrence creates a connection carrying role and Boolean polarity, with messages in both directions.
\subsection{Validation Protocol}
\label{app:validation}
The reliability tables assess the correctness outputs on the three benchmarks before subsequent LLM repairs, while human-review of FOLIO, NLCO, Zebra, and GSM8K already include those repairs.
\paragraph{Rule checks.}
SATBench compares Boolean clauses, identities, and polarity with visible conditions. Logical checks variables, ordering clues, and query candidates. Hamiltonian checks vertex coverage, distinctness, and allowable transitions. These checks derive references from the visible problem description, without gold answers or target outcomes. The other four benchmarks have partial rule checks.
\paragraph{Human review.}
Independent human reviewers check 20 source--CIR pairs per remaining benchmark. A question passes if variables, domains, constraints, and query are preserved without unstated conditions or an answer. Review measures reliability and does not trigger repairs.
\subsection{CIR Example}
The coal question in the qualitative analysis (\texttt{gsm8k\_direct:formal:1184}) produces the following CIR.
\begin{lstlisting}
{
  "cir_version": "1",
  "constraints": [
    {
      "expr": {
        "left": {
          "id": "total_kg",
          "op": "var"
        },
        "op": "eq",
        "right": {
          "op": "const",
          "value": 850
        }
      },
      "id": "total_kg_fact"
    },
    {
      "expr": {
        "left": {
          "op": "linear_sum",
          "terms": [
            {
              "coef": 50,
              "var": "bags"
            }
          ]
        },
        "op": "eq",
        "right": {
          "id": "total_kg",
          "op": "var"
        }
      },
      "id": "bag_weight_relation"
    },
    {
      "expr": {
        "left": {
          "op": "linear_sum",
          "terms": [
            {
              "coef": 18,
              "var": "bags"
            }
          ]
        },
        "op": "eq",
        "right": {
          "id": "cost",
          "op": "var"
        }
      },
      "id": "cost_relation"
    }
  ],
  "entities": [],
  "query": {
    "expr": {
      "id": "cost",
      "op": "var"
    },
    "sense": "minimize",
    "type": "optimization"
  },
  "variables": [
    {
      "domain": "nonnegative",
      "id": "total_kg",
      "meaning": "total kilograms of coal ordered",
      "type": "numeric"
    },
    {
      "domain": "nonnegative integer",
      "id": "bags",
      "meaning": "number of 50 kg bags in the order",
      "type": "numeric"
    },
    {
      "domain": "nonnegative",
      "id": "cost",
      "meaning": "total cost of the order in dollars",
      "type": "numeric"
    }
  ]
}
\end{lstlisting}

\section{Features and Training Details}
\label{app:stats}
\begin{table}[ht]
{\centering\small
\setlength{\tabcolsep}{3pt}
\begin{tabular}{p{0.16\columnwidth}p{0.65\columnwidth}r}
\toprule
Family & Features & Count\\
\midrule
Size & \raggedright Counts of all nodes, variable nodes, factor nodes,
non-query factor nodes, and edges & 5\\
\addlinespace[2pt]
Factor & \raggedright Occurrences of variables, propositions, or entities in each factor$^{*}$ & 4\\
\addlinespace[2pt]
Degree & \raggedright Degrees of variable nodes$^{*}$ (4) and factor nodes$^{*}$ (4) & 8\\
\addlinespace[2pt]
Topology & \raggedright Connected components; density; 2-core fraction;
largest biconnected-component fraction; cycle rank & 5\\
\addlinespace[2pt]
Query & \raggedright Query degree; fraction of reachable constraints;
mean distance to reachable constraints & 3\\
\addlinespace[2pt]
CIR\newline operators & \raggedright Occurrence counts of $\neg,\land,\lor,\Rightarrow,\forall,\exists,
=,\ne,<,\le,>,\ge$, linear sum, predicate, variable, and constant & 16\\
\midrule
\multicolumn{2}{l}{Total} & 41\\
\bottomrule
\end{tabular}}
\caption{\syss features, grouped by family. Variable nodes denote all non-factor nodes.}
\label{tab:stats_features}
\vspace{1mm}
\footnotesize
$^{*}$We compute mean, maximum, 90th percentile and coefficient of variation for these quantities.
\end{table}

\paragraph{Feature definitions.}
Size includes auxiliary factors; constraint nodes are non-query factors. Factor counts repeated occurrences, whereas Degree counts distinct neighbors. Summaries use mean, maximum, 90th percentile, and population SD/mean (zero when the mean is zero). Density is $2|E|/(|V|(|V|-1))$; 2-core and largest biconnected-component fractions divide by $|V|$; cycle rank is $|E|-|V|+c$ for $c$ components. We compute topology with NetworkX~\cite{hagberg2008networkx}.

Query features use the first query factor's degree, reachable non-query factor fraction, and mean distance, with missing values when undefined. Operator features recursively count occurrences, including query expressions.
\subsection{Encoder and Pretraining}
\label{app:training}
Kind, subtype and factor-type embeddings have dimensions 8, 8 and 4, with 5, 64 and 4 categories. Node subtypes record variable or entity types and
predicate argument counts. The 20 embedding coordinates and 16 log-transformed operator counts are projected from 36 to 64 dimensions with ReLU and layer normalization.

Each of the six layers has role/polarity embeddings of size $32\times64$/$4\times64$, a learned scalar $\epsilon$, a 64--128--64 MLP, ReLU, dropout 0.1, and residual layer normalization. Readout concatenates sum, mean, query sum, and population SD into 256 dimensions, using variance stabilizer $10^{-8}$.

Node/edge masking probabilities are 0.4/0.2. Learned vectors replace masked attributes; reverse edges share a mask and connectivity is retained. One mask is forced when a nonempty set receives none. Six equally weighted losses reconstruct kind, subtype, factor type, operator counts, edge role, and polarity. Subtype applies to non-factors, factor type/operators to factors, and edge losses to one orientation per masked pair. Empty losses are zero. Linear decoders use 64D node states or 128D endpoint concatenations.

We use AdamW, learning rate 0.001, weight decay 0.001, batch size 32, gradient-norm clipping 5, and fixed 100 epochs. Each epoch samples 4,684 graphs with equal source weight; validation balances 516 entries with mask seed 777001. Encoder seeds are 20260919, 20260922, 20260923, and 20260924. Evaluation freezes the encoder and disables dropout and masking.
\subsection{Gradient-Boosted Predictor}
We use scikit-learn's \texttt{HistGradientBoostingClassifier}~\cite{pedregosa2011scikit}, with the shared grid $(\text{maximum leaves},\text{iterations},\ell_2)$:
\begin{equation}
\{(3,50,1),(3,100,10),(5,50,1),(5,100,10)\}.
\end{equation}
Fixed settings are learning rate 0.05, maximum depth 3, minimum leaf size 40, early stopping disabled, and seed 713. Each question supplies binary rows 0/1 weighted by $1-y_i$/$y_i$; leaf size counts these rows.

Training-fold preprocessing performs median imputation (zero for all-missing columns), appends missingness indicators, drops constant columns, and standardizes. Inner-validation NLL selects the candidate, with grid order breaking ties; final fitting uses the outer training fold. This procedure is shared by Stats and both GNN variants.

\section{Baseline Implementations}
\label{app:baselines}
Baselines share CIRCE's evaluation labels and outer folds. We retain their published predictor families, with the settings below.
\paragraph{C0 and self probability.}
C0 predicts the training-fold mean. Self probability asks the target once for a pre-answer JSON probability in $[0,1]$, using its answering settings (Appendix~\ref{app:prompts}). Training-only calibration fits a regularized logistic intercept and slope on standardized log-odds, with regularization 0.001. Missing estimates use C0: eight for Qwen3.8-27B (SATBench 1, NLCO 6, Hamiltonian 1), zero for other targets.
\paragraph{ADeLe.}
The 18 original demand rubrics are scored by GPT-6 Sol with temperature 0, top-$p$ 1, reasoning None, and output cap unset. Unguessability is $100(1-1/K)$ for $K$ choices and 100 for open-ended tasks. A random forest uses 100 trees, entropy criterion, and minimum split size 50, with remaining defaults. Training expands each question into ten grouped binary attempt rows.
\paragraph{PredictaBoard.}
\texttt{text-embedding-3-large} embeds the visible question into 3,072 dimensions. Inner-validation NLL selects default L2 logistic regression, L1 logistic regression (liblinear, $C\in\{1,0.1\}$) or default XGBoost~\cite{chen2016xgboost} with one CPU thread. Binary rows have weights $k_i$ and $10-k_i$.
\paragraph{Already Knows.}
Qwen3.5-9B generates one thinking-off trajectory per question. Last-layer 4,096D states train a 4096--64--1 MLP with GELU and sigmoid: AdamW, learning rate $10^{-4}$, weight decay $10^{-3}$, 40 epochs and discount 1. The next-state target is detached. Terminal reward is the trajectory's response correctness. Squared TD errors are summed across the whole response, followed by one optimizer step. Chunks of 1,024 states bound memory; caches are FP16 and computations FP32.

TD training uses 199 complete NLCO trajectories and 200 per other benchmark. Evaluation uses the pre-generation state $h_0$ for all questions. Our setup uses approximately 160 training questions per fold and attempt-weighted AUROC.

\section{Compute and Software}
\label{app:compute}
CIRCE pretraining and predictors run on local CPU virtual machines with 8 cores. Each pretraining worker uses two CPU threads, while tree-fitting workers use one. The reconstruction module has 127,995 parameters, comprising 117,118 encoder, 10,429 decoder, and 448 mask parameters. Qwen targets have 9B/27B parameters. Inkling has 975B total and 41B active. Hosted GPT-6 model sizes are undisclosed.
\begin{table}[ht]
\centering\small
\setlength{\tabcolsep}{3pt}
\begin{tabular}{rrrr}
\toprule
Encoder seed & Minutes & Train loss & Val. loss\\
\midrule
20260919 & 22.39 & 0.01149 & 0.01090\\
20260922 & 22.27 & 0.01145 & 0.01003\\
20260923 & 23.18 & 0.01082 & 0.01059\\
20260924 & 22.50 & 0.01035 & 0.02221\\
\bottomrule
\end{tabular}
\caption{100-epoch pretraining: worker wall time in minutes and final training/validation reconstruction losses. Each seed uses two CPU threads.}
\label{tab:app-training-time}
\end{table}

Training through epoch 100 across four encoder seeds totals 1.51 worker wall-hours. The 336 pretrained predictor evaluations total 0.56 worker wall-hours. Across 479 recorded exploratory SSL runs, the summed worker wall time is 379.67 hours.

The encoder uses PyTorch~\cite{paszke2019pytorch}, graph features use NetworkX, and predictors use scikit-learn and XGBoost. Software versions for metric recomputation appear in Table~\ref{tab:app-software}.
\begin{table}[ht]
\centering\small
\setlength{\tabcolsep}{3pt}
\begin{tabular}{lr}
\toprule
Package & Version\\
\midrule
torch & 2.13.0+cu130\\
numpy & 2.4.3\\
scipy & 1.17.1\\
sklearn & 1.8.0\\
networkx & 3.6.1\\
xgboost & 3.4.1\\
transformers & 5.12.1\\
safetensors & 0.8.0\\
python & 3.13.12\\
\bottomrule
\end{tabular}
\caption{Software versions used for metric recomputation.}
\label{tab:app-software}
\end{table}

\section{Responsible Research}
\label{app:responsible}
\paragraph{Use and data.}
CIRCE estimates correctness under a specified model and inference protocol. Incorrect confidence can encourage over-reliance; new languages and applications require validation. We use released English reasoning benchmarks. A contact-pattern screen of 1,400 evaluation prompts found no email addresses, HTTP(S) URLs, or international phone-number matches. Names defining problem entities are retained in text and omitted from graph features.
\paragraph{Asset terms.}
Upstream licenses and sources are:
\begin{itemize}[leftmargin=*,nosep]
\item SATBench (Apache-2.0): \url{https://huggingface.co/datasets/LLM4Code/SATBench}.
\item FOLIO v2 (MIT dataset card): \url{https://huggingface.co/datasets/yale-nlp/FOLIO}; legacy repository: CC-BY-SA-4.0.
\item NLCO (MIT): \url{https://huggingface.co/datasets/summer142857jiang/NLCO}.
\item Enigmata (Apache-2.0): \url{https://github.com/BytedTsinghua-SIA/Enigmata}.
\item GSM8K (MIT): \url{https://github.com/openai/grade-school-math}.
\item BBH (MIT): \url{https://github.com/suzgunmirac/BIG-Bench-Hard}; BIG-Bench (Apache-2.0): \url{https://github.com/google/BIG-bench}. Task-specific notices apply.
\end{itemize}

\section{Prompt Templates}
\label{app:prompts}
\subsection{General CIR Translation}
\begin{lstlisting}
You convert natural-language structured reasoning problems into CIR v1 JSON.

Your job is REPRESENTATION ONLY. Do not solve the problem, do not choose an
answer option, do not derive the final assignment, and do not use any hidden or
gold information.

CIR v1 has exactly five root fields:
- cir_version: "1"
- entities: list of {"id": string, "type": string, "label": string}
- variables: unknown Boolean/numeric/assignment decisions. Each has id and type,
  and may have meaning/domain/bounds.
- constraints: list of {"id": string, "expr": EXPR}
- query: one of
  {"type":"feasibility"}
  {"type":"entailment","expr":EXPR}
  {"type":"optimization","sense":"minimize"|"maximize","expr":EXPR}

Registered expressions include:
- variable/entity/constant:
  {"op":"var","id":"..."}
  {"op":"entity","id":"..."}
  {"op":"const","value":...}
- Boolean:
  {"op":"not","arg":EXPR}
  {"op":"and","args":[...]}
  {"op":"or","args":[...]}
  {"op":"xor","args":[...]}
  {"op":"implies","left":EXPR,"right":EXPR}
  {"op":"iff","left":EXPR,"right":EXPR}
- comparisons:
  {"op":"eq"|"ne"|"le"|"lt"|"ge"|"gt","left":EXPR,"right":EXPR}
- linear sums:
  {"op":"linear_sum","terms":[{"coef":1,"var":"x"}, ...]}
- predicates:
  {"op":"predicate","name":"Relation","args":["entity_or_bound_var", ...]}
- quantifiers:
  {"op":"forall"|"exists","vars":[{"id":"x","type":"entity"}],"body":EXPR}
- scheduling relations when explicitly appropriate:
  precedence, no_overlap, within_window, cumulative_resource.

Representation rules:
1. Use ONLY the problem text below.
2. Preserve all stated constraints that determine the latent reasoning world.
3. Do not invent unstated facts or commonsense rules unless the text explicitly
   states that they are task rules.
4. For ordering/allocation/grid/full-assignment puzzles, encode the latent
   decision variables and constraints and use a feasibility query. The task of
   finding the assignment is represented by the feasible solution space; do not
   put the answer assignment in CIR.
5. If the question asks whether a proposition follows and that proposition can
   be expressed without solving, use an entailment query.
6. If the visible problem is multiple choice or asks for a selected value but
   CIR v1 cannot faithfully express the selection query without solving, encode
   the underlying world constraints and use feasibility. Never encode the
   correct choice as a constraint.
7. Answer choices are not facts. They may help identify what relation the query
   refers to, but they must not be asserted as constraints.
8. For truth-teller/liar puzzles, represent each person's identity as an
   unknown Boolean decision and encode the semantic relationship between the
   speaker's identity and the truth value of their stated proposition.
9. For logic-grid puzzles, use generic assignment/position decisions and
   all-different/exactly-one structure expressed with Boolean variables and
   linear sums where practical.
10. Output exactly one CIR JSON object after any thinking. No Markdown and no
    answer/solution/commentary fields.

PROBLEM TEXT:
{visible_text}

Produce CIR v1 JSON only.
\end{lstlisting}
\subsection{Format-Only Repair}
\begin{lstlisting}
Reformat the raw model output below as exactly one valid CIR v1 JSON
object. Preserve only the representation already stated in the raw output. Do
not solve the task, choose an answer, add/remove a condition, or consult any
outside input. Return no Markdown.

Exactly these root fields are allowed:
{"cir_version":"1","entities":[],"variables":[],"constraints":[],"query":{}}

RAW OUTPUT:
\end{lstlisting}
\subsection{Pre-Answer Self Probability}
\begin{lstlisting}
Estimate your own probability of producing a correct final answer to the task below on ONE independent attempt, following the task's original instructions and the answering settings listed below.

Do not answer the task, provide a candidate answer, or work out its solution. Assess the task requirements and your own likely performance before attempting an answer. The instructions inside <task> describe the hypothetical answering task; do not execute them now.

Answering settings: {settings}

<task>
{question}
</task>

Return only one JSON object in this format:
{{"probability_correct": <number between 0 and 1>}}
This is the probability that your final answer would be correct on one attempt, not the probability of being correct on every attempt. Do not include an answer, explanation, or any other fields.
\end{lstlisting}

\end{document}